%% file: main.tex
\documentclass[sigconf,9pt]{acmart}
\renewcommand\footnotetextcopyrightpermission[1]{} 
\usepackage{algorithm}
\usepackage{amsmath}
\usepackage{xcolor}
\usepackage{float}
\usepackage{graphicx}
\usepackage{listings}
\usepackage{mathrsfs}
\usepackage{multirow}
\usepackage{multicol}
\usepackage{pifont}
\usepackage{siunitx}
\usepackage{subfigure}
\usepackage{subfloat}

\usepackage{url}

\usepackage{algpseudocode}
\algtext*{EndWhile}
\algtext*{EndIf}
\algtext*{EndFor}
\algtext*{EndProcedure}

\usepackage{colortbl}
\usepackage{dcolumn}

\definecolor{greenDeep}{RGB}{0,170,0}
\definecolor{greenShallow}{RGB}{0,255,0}
\definecolor{greenShallower}{RGB}{160,255,0}
\definecolor{orangeShallow}{RGB}{255,190,0}
\definecolor{orangeDeep}{RGB}{255,80,0}
\definecolor{orangeDeeper}{RGB}{255,40,0}
\definecolor{redDeep}{RGB}{255,0,0}

\def\zz#1{%
\ifdim#1pt>0.85pt\cellcolor{greenDeep}\else
\ifdim#1pt>0.789pt\cellcolor{greenShallow}\else
\ifdim#1pt>0.73pt\cellcolor{greenShallower}\else
\ifdim#1pt>0.67pt\cellcolor{yellow}\else
\ifdim#1pt>0.62pt\cellcolor{orangeShallow}\else
\ifdim#1pt>0.57pt\cellcolor{orange}\else
\ifdim#1pt>0.52pt\cellcolor{orangeDeep}\else
\ifdim#1pt>0.45pt\cellcolor{orangeDeeper}\else
\cellcolor{redDeep}\fi\fi\fi\fi\fi\fi\fi\fi
#1}

\newcommand{\thistheoremname}{}
\newtheorem*{genericthm*}{\thistheoremname}
\newenvironment{namedthm*}[1]
  {\renewcommand{\thistheoremname}{#1}%
   \begin{genericthm*}}
  {\end{genericthm*}}

\newtheorem*{problem*}{Problem}

\usepackage{mathtools}

\usepackage{url}

\graphicspath{{./_fig/}}

\usepackage{amsthm}

\makeatletter
\def\@copyrightspace{\relax}
\makeatother
\begin{document}


\title{\vspace{-.1in}Towards Collaborative Intelligence: Routability Estimation \\based on Decentralized Private Data}

\author{\large{Jingyu~Pan$^1$, Chen-Chia~Chang$^1$, Zhiyao~Xie$^2$, Ang~Li$^1$, Minxue~Tang$^1$, Tunhou~Zhang$^1$, Jiang~Hu$^3$, and Yiran~Chen$^1$}}
\affiliation{%
 \vspace{-.1in}
 \institution{Duke University$^1$ \hspace{0.3em} Hong Kong University of Science and Technology$^2$ \hspace{0.3em} Texas A\&M University$^3$}
}
\affiliation{%
  \institution{\small{\{jingyu.pan, chenchia.chang, ang.li630, minxue.tang, tunhou.zhang, yiran.chen\}@duke.edu, eezhiyao@ust.hk, jianghu@tamu.edu}}
}

\input{_txt/abstract}



\maketitle

\vspace{-.1in}
\input{_txt/1_introduction}

\input{_txt/2_preliminaries}

\input{_txt/3_Problem_Formulation}

\input{_txt/4_Methodology}

\input{_txt/5_experiment}

\input{_txt/6_conclusion}

\input{_txt/7_acknowledgements}


\bibliographystyle{ACM-Reference-Format}
\bibliography{references}
\end{document}

%% file: _txt/abstract.tex
\begin{abstract}

Applying machine learning (ML) in design flow is a popular trend in Electronic Design Automation (EDA) with various applications from design quality predictions to optimizations.
Despite its promise, which has been demonstrated in both academic researches and industrial tools, its effectiveness largely hinges on the availability of a large amount of high-quality training data.
In reality, EDA developers have very limited access to the latest design data, which is owned by design companies and mostly confidential.
Although one can commission ML model training to a design company, the data of a single company might be still inadequate or biased, especially for small companies.
Such data availability problem is becoming the limiting constraint on future growth of ML for chip design. 
In this work, we propose an Federated-Learning based approach for well-studied ML applications in EDA.
Our approach allows an ML model to be collaboratively trained with data from multiple clients but without explicit access to the data for respecting their data privacy.
To further strengthen the results, we co-design a customized ML model FLNet and its personalization under the decentralized training scenario.
Experiments on a comprehensive dataset show that collaborative training improves accuracy by 11\% compared with individual local models, and our customized model FLNet significantly outperforms the best of previous routability estimators in this collaborative training flow.

\end{abstract}


%% file: _txt/1_introduction.tex
\section{Introduction}
EDA techniques have achieved remarkable progress over past decades. 
However, the current chip design flow is still largely restricted to individual point tools with limited interplay across different tools and design steps.
Tools in early steps cannot well judge if their solutions may eventually lead to satisfactory designs, and the consequence of a poor solution cannot be found until very late.
Such disjointedness in the design flow is traditionally mitigated by either simplified estimations with heuristics or iterative design, which often lead to over-conservative design or longer turn-around time, respectively.
To improve the predictability in chip design flow, ML models have been constructed based on prior data to provide early feedback or help accelerate the solving of EDA problems.  

In recent years, ML for EDA has become a trending topic~\cite{huang2021machine}.
ML models are applied at almost all design stages of the VLSI design flow, including high-level synthesis, logic synthesis, and physical design~\cite{xie2022pre, xie2018routenet, yu2019painting, chen2020pros}, making predictions on timing~\cite{xie2022pre}, power~\cite{xie2021apollo}, routability~\cite{xie2018routenet, yu2019painting, chen2020pros}, etc.
These ML models learn from prior solutions and typically provide orders-of-magnitude faster design quality evaluations.
Besides being a hot research topic in academia, ML-based estimators have also gained popularity in the EDA industry.
Recent versions of commercial tools already support the construction of ML models on delay~\cite{Innovus} or congestion predictions~\cite{ICC2}.
The vendors also claim improved PPA or faster convergence after invoking the ML models in their tools~\cite{Innovus, ICC2}.
In summary, ML for EDA has demonstrated its impressive contribution to the quality of result (QoR) and overall turn-around time in both academia and industry.

However, despite the proven advantages, there are still obstacles that prevent wide applications of ML models in EDA.
One vital but rarely explicitly discussed challenge is the availability of data.
The data includes circuit designs and corresponding chip qualities including power, performance, and etc.
Most latest design data are owned by design companies and are highly confidential. 
As a result, many works from academia have to construct models with designs from public benchmarks, most of which are either outdated or over-simplified only for contest purposes. 
Data availability also affects the quality of practical ML for EDA models in industry.
Although one can train ML models within a design company, the data of a single company might still be inadequate or biased, especially for small companies.
As for EDA vendors, due to data privacy concerns, some commercial tools~\cite{Innovus} have to construct ML models from scratch, only using the limited raw data provided by each client. 
In summary, lack of data has seriously hampered the adoption of ML models in EDA, yet very limited research explorations can be found for mitigating this problem. 

To address the above challenges and promote collaborative intelligence, we present a novel framework to collaboratively train ML models without explicitly collecting or viewing data from decentralized design owners.
This framework is demonstrated on routability estimation, which is a well-studied problem in ML for EDA.
To promote collaborative learning, we co-design the federated learning ML model architecture, named FLNet, and its personalization techniques to seek unprecedented solutions in federated learning.
The proposed approach, when put in practice, allows design companies to leave private data on their own servers and admit only pre-determined operations and communications\footnote{Engineering details including the implementation of federated learning framework~\cite{beutel2020flower, hu2019fdml} and privacy concerns~\cite{naseri2020toward, sun2020provable} about the framework on both model and data have been well studied for general machine learning tasks. These engineering details are not special in ML for EDA, thus are not the focus of this paper.}.
In this way, design companies can jointly utilize their design data for ML model construction without disclosure of their product.
For example, EDA vendors can first train a significantly more general ML model using designs from all its cooperators.
Then each cooperator, as a client, can optionally customize the general model for itself by leveraging personalization techniques (e.g. local fine-tuning) using its own private data to achieve better performance.

Our contributions in this work are summarized as follows.

\begin{itemize}
\item We bring attention to the data availability problem in ML for EDA, and propose a collaborative training solution to encourage data sharing while respecting data privacy.  
\item We co-design the personalization flow and network architecture for our proposed decentralized training scenario. With the best personalization, FLNet outperforms previous routability estimators by 11\% in accuracy\footnote{This framework is open-sourced in \url{https://github.com/panjingyu/Decentralized-Routability-Estimation}.}.
\item 
We provide detailed ablation studies to justify our co-design on personalization and network architectures.
\end{itemize}

%% file: _txt/2_preliminaries.tex
\section{Preliminaries}

\subsection{Routability Estimation}

We demonstrate our algorithm on routability estimation, since it is a representative and well-studied topic~\cite{xie2018routenet, yu2019painting, liang2020drc, chen2020pros} in ML for EDA, and its problem formulation and solutions share many similarities with other important EDA problems like IR drop estimation~\cite{fang2018machine}, clock tree prediction~\cite{lu2019gan}, lithography hotspot detection~\cite{jiang2020efficient, yang2017imbalance}, optical proximity correction (OPC)~\cite{yang2019gan}, etc.
Previous works on these problems typically borrow ideas from computer vision and adopt deep learning techniques including convolutional neural network (CNN) or generative adversarial network (GAN) to process the features from circuit layouts. 

Previous routability estimators use either routing congestions~\cite{chen2020pros, yu2019painting} or DRC (design rule checking)~\cite{xie2018routenet, liang2020drc} as the metric of routability.
They detect congestion locations or DRC hotspots.
Given a set of placement solutions with extracted input feature maps $X_i$, routability estimators generate a neural network model $f$ to detect the locations of DRC hotspots or congestions $Y_i$:
$$
f: X_i\in \mathbb{R}^{w\times h\times c} \rightarrow Y_i
\in \{ 0, 1 \}^{w \times h}
$$
where $w$ and $h$ are the width and height of the layout, and $c$ indicates the number of input features/channels. 

We demonstrate our learning algorithm through comparisons with two representative routability estimators from RouteNet~\cite{xie2018routenet} and PROS~\cite{chen2020pros}.
Both works adopt fully convolutional network (FCN)-based estimators with significantly different model structures.
The estimator from RouteNet~\cite{xie2018routenet}, as an earlier work, consists of only convolution layers, trans-convolutional layers and a shortcut structure.
In comparison, the estimator from PROS~\cite{chen2020pros} adopts more advanced structures including dilated convolution~\cite{yu2015multi} blocks, refinement blocks, and sub-pixel upsampling blocks.

\subsection{Federated Learning}

Federated learning (FL) includes a series of decentralized training techniques, proposed for their distinct privacy protection advantage compared with training on a central machine with persistent data~\cite{mcmahan2017communication}.
FedAvg~\cite{mcmahan2017communication} is a popular FL algorithm for most computer vision tasks.
In FedAvg, the decentralized training is performed iteratively.
In each round, the clients send updates of locally trained models to the central server, and the server then averages the collected updates and distributes the aggregated update back to all the clients.
FedAvg works well with independent and identically distributed (IID) datasets but may suffer from significant performance degradation when it is applied to non-IID datasets.



%% file: _txt/3_Problem_Formulation.tex
\section{Problem Formulation}

We assume there are altogether $K$ clients providing their data for model training.
In practice, the circuit designs from the same client/company tend to be more similar to each other, since they may be from the same series of products, while different clients may provide largely different designs.
This is reflected in our experiment by only assigning designs from the same benchmark to the same client.
Assume each client $k \in [1, K]$ provides $n_k$ data samples.
For the routability estimation task, each data sample includes one placement solution, and its label is the ground-truth DRC hotspot map.
The training data for each client $k$ is denoted as $\{X_i, Y_i\}_k \  (i \in [1, n_k], k \in [1, K])$, where $X_i \in \mathbb{R}^{w\times h\times c}$ is the feature map of a placement with $w \times h$ grids and $c$ channels, while $Y_i \in \{0, 1\}^{w \times h}$ is the hotspot distribution.
To verify the estimator accuracy, each client also has their own testing data $\{X^{Test}_i, Y^{Test}_i\}_k \  (k \in [1, K])$, which is generated from circuit designs completely different from those of the training data.
Traditionally, if the developer directly commission ML model training to each client, it results in $K$ local models trained only by each client's local data.
These models are baselines denoted as $b_k$, which is trained on $\{X_i, Y_i\}_k$ by the client $k$.

A key difference in our problem setting compared with previous works is the \textit{data privacy} constraint.
All clients' data should be private to themselves, which means no party other than the client itself should have the access to its training and testing data.
Under the data privacy constraint, there are two goals in this work.
First, we develop a general model that achieves higher performance on all $K$ clients' testing data $\{X^{\text{Test}}_i, Y^{\text{Test}}_i\}_k (k \in [1, K])$.
This model should generalize better than local model baselines.
Second, we build a customized model for each client for better local accuracy. 



%% file: _txt/4_Methodology.tex
\section{Methodology} \label{sec:algorithm}




In this work, we try to train our model for routability estimation utilizing data from all $K$ clients without violating their data privacy.
Compared with baselines trained on each local client, we enlarge the training dataset by $K$ times.
To achieve this, we need a strict decentralized training setting with limited data access for privacy protection.
And the decentralized training setting poses challenges to successful model construction.
The client-level data heterogeneity commonly seen in routability estimation tasks makes decentralized training suffers from convergence issues and performance degradation.
To make thing worse, existing models are too complex and involve special operators that are not robust against some operations in the decentralized training setting, thus failing to perform well.
Therefore, in the following sections, we 1) analyze the current challenges and a solution (FedProx) of decentralized training for routability prediction;
2) propose \textbf{FLNet}, a novel model customized for better performance in the limited decentralized training setting;
3) explore \textbf{model personalization techniques} to alleviate the negative impact of client-level data heterogeneity;
4) briefly mention the features we use for routability estimation.

\subsection{When Decentralized Training meets Routability Estimation}

\begin{figure}[tb]
    \centering
    \includegraphics[width=0.92\linewidth]{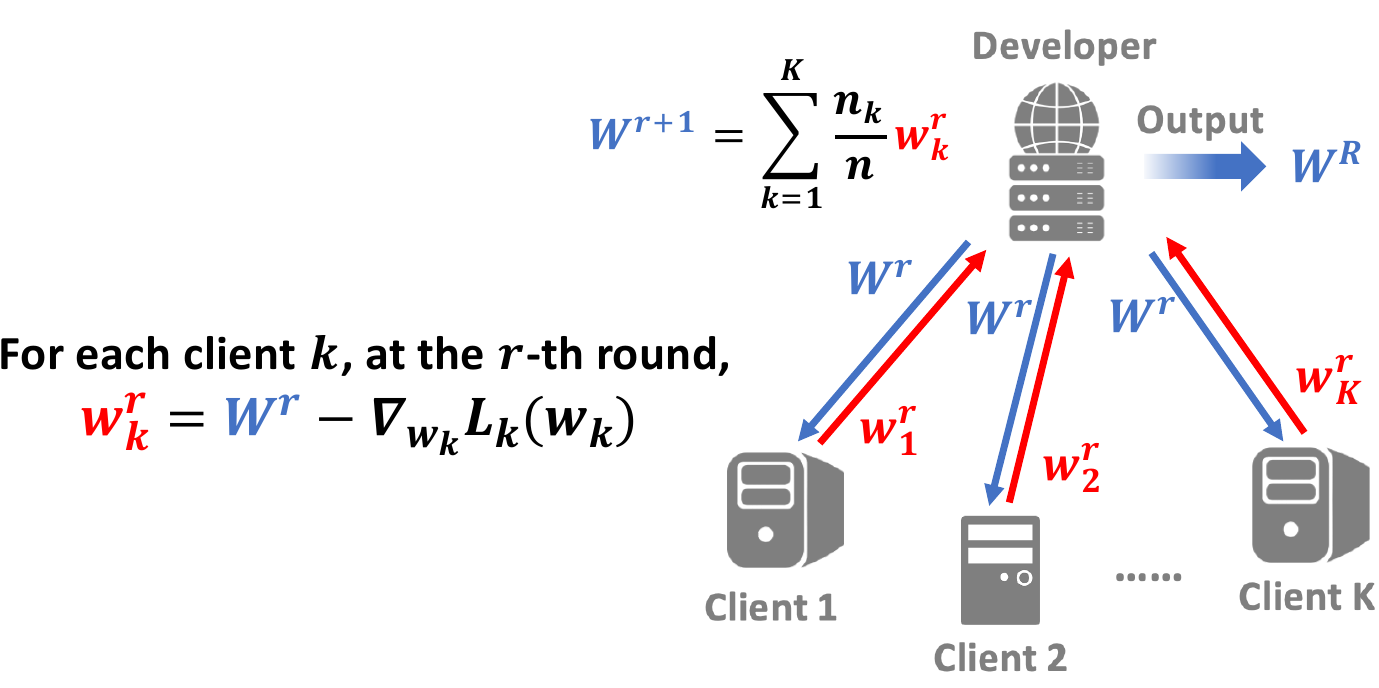}
    \vspace{-.15in}
    \caption{The visualization of the decentralized training setting. $R$ denotes the total number of rounds. $K$ denotes the number of clients. $w$ and $W$ denotes the locally trained models and the aggregated models, respectively.
    }
    \label{fig:decentralized}
\end{figure}

The decentralized training setting requires all training and optimization operations to take place only at the client side, without gathering data from any client.
And the developer can only receive model parameters from its clients, perform parameter aggregation, and deploy the average parameters back to the clients.
Figure~\ref{fig:decentralized} shows the visualization of the decentralized training setting with $K$ clients.
For each round $r \in [1, R]$, each client $k$ trains the model on its own data, and send the trained parameters $w_k^r$ to the developer.
Then the developer performs parameter aggregation on all collected parameters and generate an average model $W^{r+1} = \sum_{k=1}^K \frac{n_k}{n} w_k^r$.
To conclude round $r$, the average model $W^{r+1}$ is deployed back to all the clients for further training at the next round.
This procedure is repeated for $R$ times and the developer will construct a generalized model $W^R$, which is the aggregated model at the $R$-th round.

\newpage
This setting brings at least two new challenges.
First, \textbf{high data heterogeneity among clients hinders the convergence of decentralized training}.
Different clients can contain largely different circuit designs in terms of functionality or microarchitecture.
The intrinsic differences between circuit designs cause the data heterogeneity in their feature distributions.
And ML models trained on different clients tend to capture non-general patterns which reflect such data heterogeneity.
The typical client-level heterogeneity of routability data makes decentralized training suffer from slow convergence and low accuracy, or even fail to converge.

Second, \textbf{existing routability estimator models do not cooperate well with decentralized training}, leading to a large performance degradation.
Existing works typically utilize complex models with very high non-linearity, and thus are over-sensitive to operations like parameter aggregation, which happens frequently in the decentralized training setting.
Besides, they also involve model components that increase convergence difficulty in the decentralized training setting.

We apply the FedProx~\cite{li2018federated} method to address the convergence issue arising from heterogeneous training data distribution among clients.
In FedProx, each client performs training on its local data, and sends the trained model parameters instead of the data to the central machine of the model developer.
The model developer then performs parameter aggregation based on the collected parameters from all clients, and sends back the average parameters to the clients for further training.
Such procedure is repeated until the model converges.
FedProx's training setting satisfies the requirements of the aforementioned decentralized training scenarios.
Furthermore, to alleviate the data heterogeneity challenge, it adds an extra proximal term to penalize the difference between each local model and the model received from the developer at the beginning of each training round.
This makes FedProx applicable to EDA tasks such as routability estimation, where heterogeneous data is very common due to huge difference in circuits designs.
In FedProx, at the $r$-th round, client $k$ optimizes the following objective $L_{\text{Prox}}$:
\vspace{-.05in}
\begin{equation}
    \min_{w_k} L_{\text{Prox}}(w_k, W^r) = \sum_{i=1}^{n_k} (w_k(X_i) - Y_i)^2 + \mu ||W^r - w_k||^2,
    \label{eq:dynmaic}
\end{equation}
where $\mu$ is a hyper-parameter that controls the contribution of the proximal term.
In this way, the difference between the global model $W^r$ and local model $w_k$ is constrained during the local training, thus preventing the divergence of local models.

\subsection{FLNet: Routability Model Customized for FL}

Existing works typically fail to adapt to the decentralized training setting because they utilize complex models that have much higher non-linearity than simple models.
Such high non-linearity introduces low robustness against fluctuation of model parameters.
And in the decentralized training setting, the model parameter fluctuation frequently happens at the parameter aggregation operation.
This can lead to much lower performance of the model compared with the same model trained in a centralized setting.
On the other hand, existing models commonly come with a large number of sequential layers, and thus need special operators like Batch Normalization layers.
Batch Normalization layers aim to ensure good convergence by whitening the input of each layer using recorded mean and variance during training.
However, when training a model with Batch Normalization layers using decentralized training setting, Batch Normalization layers usually fail to obtain stable records of mean and variance due to the frequent model parameter aggregation operation, which makes the model even harder to converge.
As a result, existing models for routability estimation generally fail to adapt well to the limited decentralized training setting and show large performance degradation.

To achieve the best performance in federated learning, we co-designed FLNet to reduce the performance degradation commonly seen in existing models for routablity estimation.
FLNet is a 2-layer CNN model without any Batch Normalization operators.
It has much fewer model parameters than previous works and thus is more robust to the negative impact from parameter fluctuation introduced by parameter aggregation.
Particularly, this parameter fluctuation is amplified by the client-level data heterogeneity commonly seen in routability data.
FLNet's robustness can protect its performance from unacceptable degradation, making it outperform existing complex models when the data is highly heterogeneous.
Table~\ref{tab:flnet} shows detailed configuration of FLNet.
Despite its simple structure, we select a large kernel size $9 \times 9$ for both layers to ensure a large receptive field at the output.
Therefore, FLNet can still capture features with a relatively large spatial range, which is important in routability estimation.

\begin{table}[tb]
    \small
    \centering
    \vspace{-.2in}
    \caption{FLNet Model Architecture Configuration}
    \vspace{-.1in}
    \resizebox{0.8\linewidth}{!}{
    \begin{tabular}{|c|c|c|c|}
        \hline
        Layer           & Kernel size   & \#Filters     & Activation \\
        \hline
        input\_conv     & 9 $\times$ 9  & 64            & ReLU       \\
        output\_conv    & 9 $\times$ 9  & 1             & None       \\
        \hline
    \end{tabular}
    }
    \label{tab:flnet}
\end{table}

\begin{figure*}[!tb]
    \vspace{-.1in}
    \subfigure[]{\includegraphics[height=0.22\textwidth]{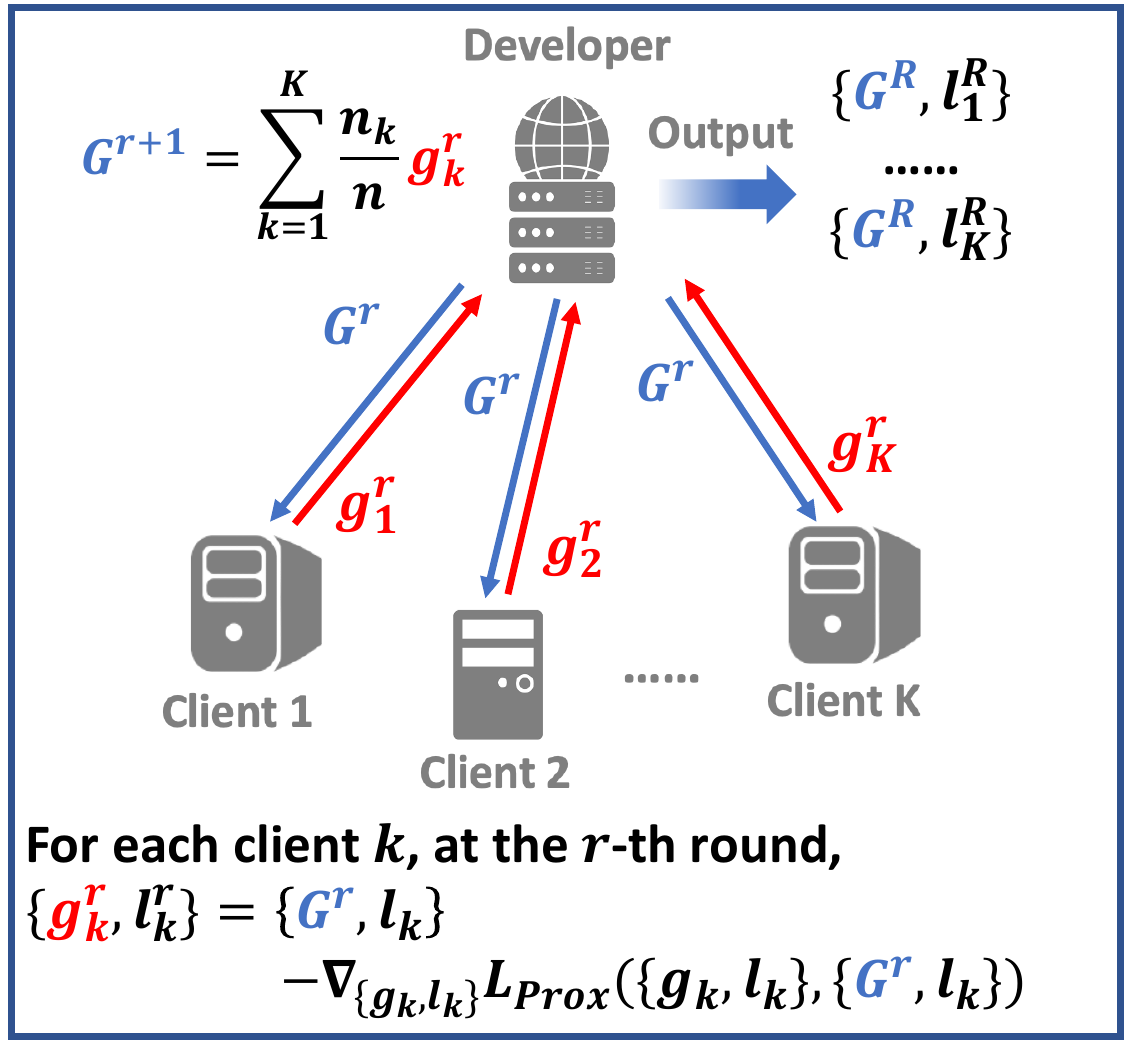}
    \label{fig:v_lg}}
    \subfigure[]{\includegraphics[height=0.22\textwidth]{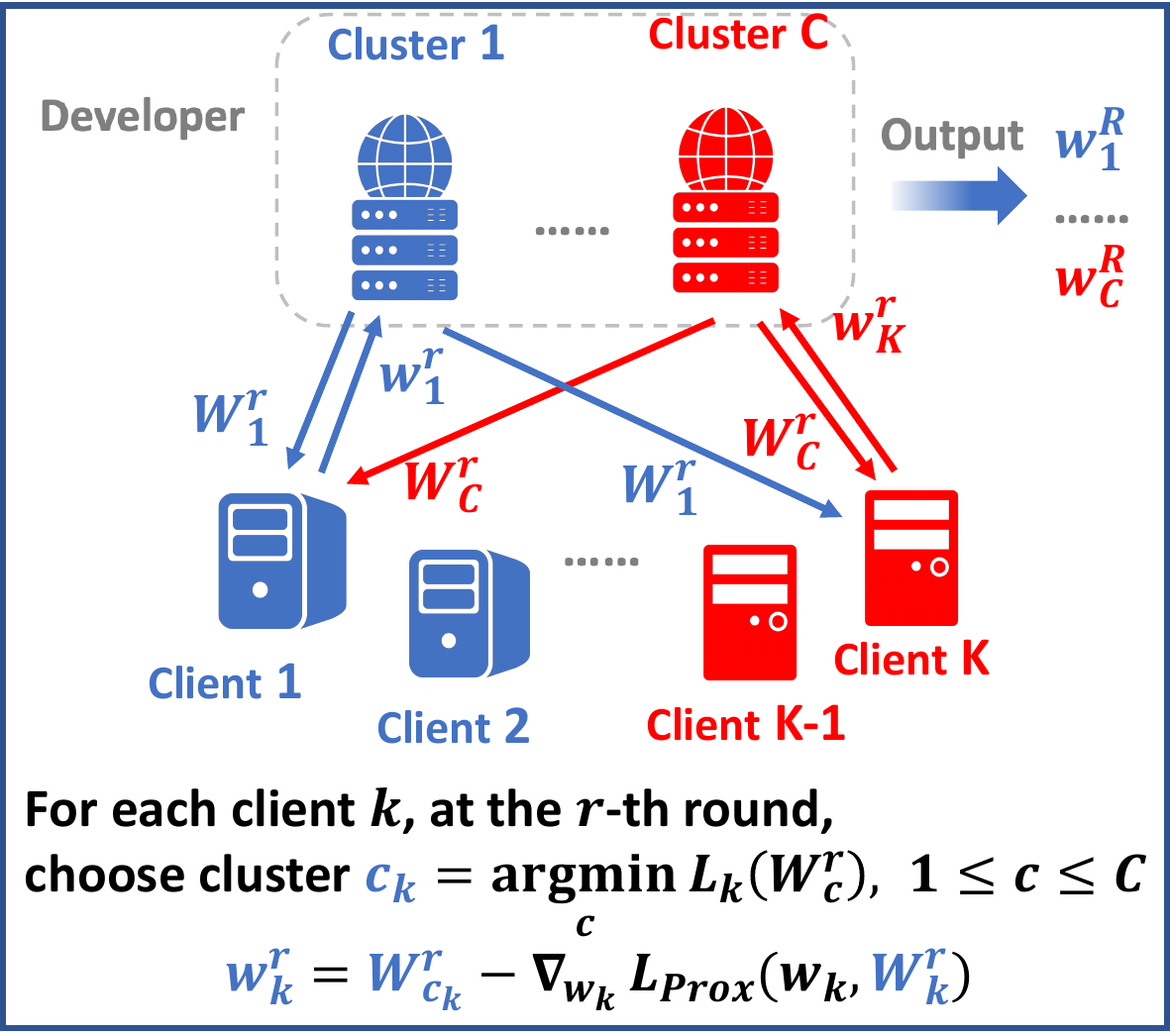}
    \label{fig:v_ifca}}
    \subfigure[]{\includegraphics[height=0.22\textwidth]{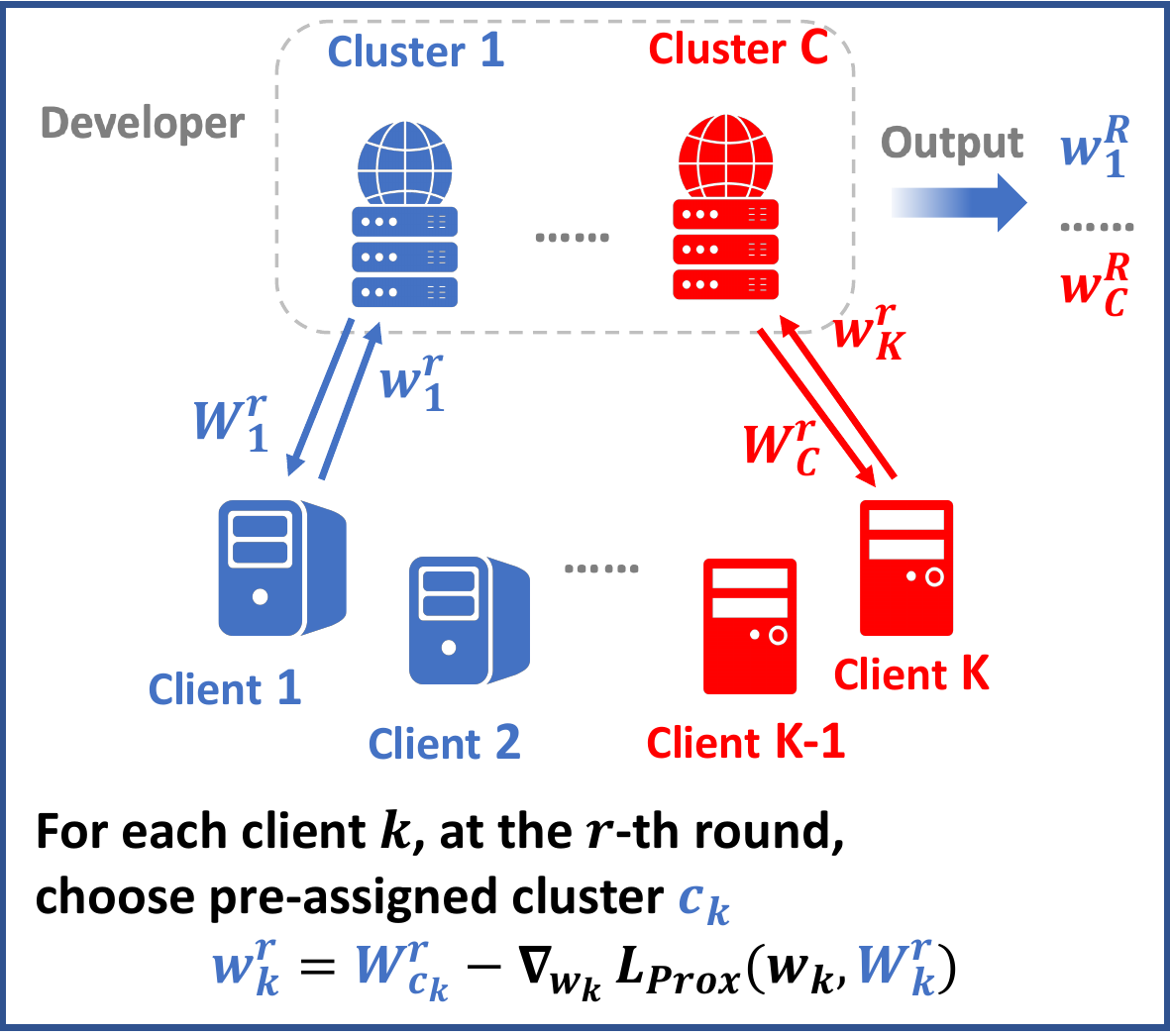}
    \label{fig:v_assigned}}
    \subfigure[]{\includegraphics[height=0.22\textwidth]{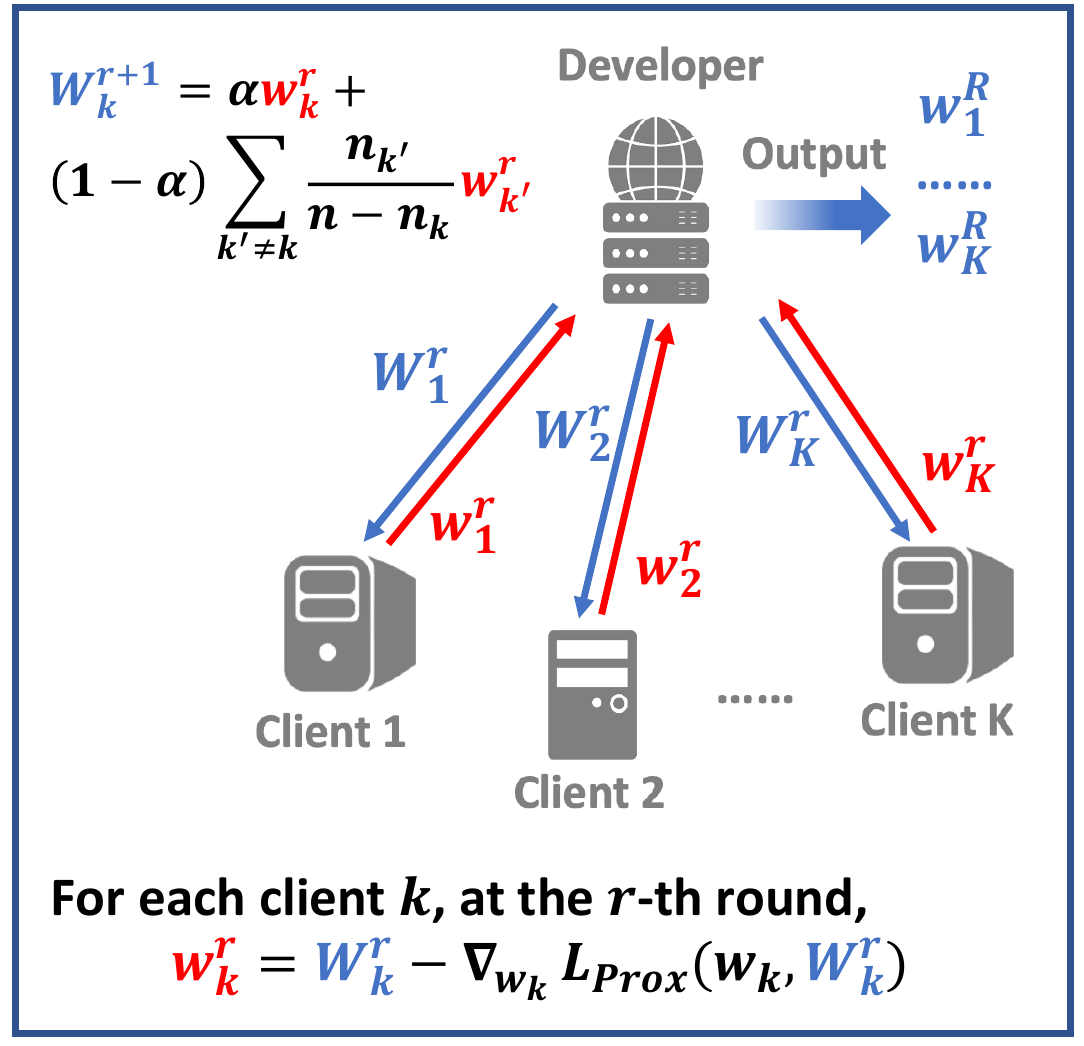}
    \label{fig:v_alpha}}
\vspace{-.2in}
\caption{Visualizations of federated learning personalization techniques based on FedProx. $g$ and $l$ denote locally trained global parts and local parts, respectively. $G$ denotes aggregated global parts. $C$ denotes the number of clusters. (a) FedProx-LG. (b) IFCA. (c) Assigned clustering. (d) FedProx + $\alpha$-portion sync.}
    \vspace{-.1in}
\end{figure*}

\subsection{Personalization in Federated Learning}

Federated learning methods (e.g., FedProx~\cite{li2018federated}) train one generalized model to achieve good average accuracy on all clients.
But personalization techniques in federated learning aim to train models that can perform better than the generalized model for individual clients that are highly data and system heterogeneous.
Typically, design companies care more about a model's accuracy on the local data than its transferability or generality.
Actually, we can further utilize model personalization techniques to trade off extra training cost and model generality for improvement of local accuracy.

In this section, we explore a set of federated learning personalization techniques based on the FedProx training scheme.
We first explore pre-existing techniques (e.g., FedProx-LG~\cite{liang2020think} and Iterative Federated Clustering Algorithm~\cite{ghosh2020efficient}) as follows:

\noindent \textbf{FedProx-LG} is based on the prior idea of \cite{liang2020think}.
Figure~\ref{fig:v_lg} shows the visualization of FedProx-LG.
FedProx-LG partitions a model into a global part $g$ and a local part $l$.
At the $r$-th round, the model developer only communicate and aggregates the global part $g^r$, leaving the local part $l^r_k, (1 \leq k \leq K)$ private to each client $k$.

\noindent \textbf{Iterative Federated Clustering Algorithm (IFCA)} introduces client-level clustering to alleviate the negative effect of data heterogeneity.
Figure~\ref{fig:v_ifca} shows the visualization of IFCA.
The $K$ clients can be categorized as $C$ clusters, and the clients of the same cluster have higher similarity and more shareable features to be captured by the DNN model.
In IFCA, the model developer initializes a model for each cluster respectively. 
At each round $r$, client $k$ determines its cluster $c_k$ by verifying the loss $L_k(W^r_c)$ of $C$ cluster models ($c = 1, 2, \dots, C$) on client $k$'s training data and chooses the cluster $c$ with the lowest loss.
Then, client $k$ trains the chosen cluster model $W^r_{c_k}$ and sends the updated model $w^r_k$ to the developer.
At the developer side, each cluster only receives and aggregates the models from its corresponding clients at that round.
The procedure repeats and the model clustering is updated iteratively.

Besides existing personalization techniques, we make our exploration comprehensive by investigating additional techniques such as local fine-tuning, assigned clustering, and $\alpha$-portion sync to further enhance FL personalization, demonstrated as follows:

\noindent \textbf{Local fine-tuning} is a simple but effective technique for model personalization.
Based on the model trained with FedProx, each client can further fine-tune the model received from the developer for extra steps on its own data.
In this way, all clients make the collaboratively-trained model adapt towards the distribution of their respective local data.

\noindent \textbf{Assigned clustering} pre-assigns a cluster for each client by leveraging prior knowledge about the clients' similarity.
Figure~\ref{fig:v_assigned} shows the visualization of the assigned clustering method.
Compared with IFCA, this method assigns a fixed cluster $c_k$ to each client $k$ for all rounds, and directly gains faster convergence and potentially higher accuracy.

\noindent \textbf{$\alpha$-portion sync} is another personalization technique with much less extra cost, where the model developer simply performs different weighted aggregation for model parameters from each client.
Figure~\ref{fig:v_alpha} shows the visualization of the $\alpha$-portion sync algorithm.
At the beginning of each round $r$, for each client $k$, the developer aggregates a customized model $W^r_k$ that takes client $k$'s previous parameters $w^{r-1}_k$ as $\alpha$-portion in the aggregation.
Compared with FedProx, the $\alpha$-portion sync method puts higher weight to each client's own parameters in the parameter aggregation.
And thus, it customizes the model to adapt more to the local data, while also learning from data of other clients.

\subsection{Feature Extraction}

Regarding the features for routability estimation, we follow previous works~\cite{xie2018routenet,chen2020pros,chang2021auto} to select features and perform feature extraction.
More specifically, our selected features capture both cell density features (e.g., locations of cells) and wire density features (e.g., connectivity between instances).
Cell density features include the routing blockage information.
Wire density features encode the instance connectivity and routing congestion information using several heuristics, such as RUDY~\cite{xie2018routenet} and fly lines.

%% file: _txt/5_experiment.tex
\section{Experiment Results}

\begin{table}[!b]
  \vspace{.15in}
  \centering
  \caption{Experiment Data Setup for Each Client}
       \renewcommand{\arraystretch}{1.2}
  \label{tbl:designs}
  \resizebox{\linewidth}{!}{
  \begin{tabular}{| l | c c c | c c c |}
 	\hline
 	\multirow{2}{*}{Clients}  &  \multicolumn{3}{c|}{ Training Designs  }  & 
 	 \multicolumn{3}{c|}{Testing Designs } \\
 	  &   \multicolumn{3}{c|}{(Num of Placements) }  &   \multicolumn{3}{c|}{(Num of Placements) }   \\
	\hline
	Client 1 &   \multicolumn{3}{c|}{4 designs in ITC'99 (462)}  & \multicolumn{3}{c|}{2 designs in ITC'99 (230)} \\
	\hline
	Client 2 &   \multicolumn{3}{c|}{2 designs in ITC'99 (231)}  & \multicolumn{3}{c|}{1 design in ITC'99 (114)} \\
	\hline
	Client 3 &   \multicolumn{3}{c|}{2 designs in ITC'99 (231)}  & \multicolumn{3}{c|}{2 designs in ITC'99 (232)} \\
	\hline
	Client 4 &   \multicolumn{3}{c|}{7 designs in ISCAS'89 (812)}  & \multicolumn{3}{c|}{3 designs in ISCAS'89 (348)} \\
	\hline
	Client 5 &   \multicolumn{3}{c|}{7 designs in ISCAS'89 (812)}  & \multicolumn{3}{c|}{3 designs in ISCAS'89 (348)} \\
    \hline
	Client 6 &   \multicolumn{3}{c|}{6 designs in ISCAS'89 (697)}  & \multicolumn{3}{c|}{3 designs in ISCAS'89 (348)} \\
    \hline
	Client 7 &   \multicolumn{3}{c|}{6 designs in IWLS'05 (656)}  & \multicolumn{3}{c|}{3 designs in IWLS'05 (280)} \\
	\hline
	Client 8 &   \multicolumn{3}{c|}{7 designs in IWLS'05 (742)}  & \multicolumn{3}{c|}{3 designs in IWLS'05 (329)} \\
    \hline
	Client 9 &   \multicolumn{3}{c|}{9 designs in ISPD'15 (175)} &  \multicolumn{3}{c|}{4 designs in ISPD'15 (84)}  \\
	\hline
  \end{tabular}
  }
\end{table}

\begin{table*}[ht]
    \centering
     \vspace{-.05in}
     \renewcommand{\arraystretch}{1.1}
     \vspace{-.1in}
     \caption{Testing Accuracy Comparison (ROC AUC) on Routability Prediction with FLNet}
     \vspace{-.1in}
     \label{result_flnet}
    \resizebox{\linewidth}{!}{
    \begin{tabular}{| c | c | c | c | c | c | c | c | c | c | c || c | }
        \hline
        \multicolumn{2}{|c|}{}   & \multicolumn{10}{c|}{Testing on}  \\
        \cline{3-12}
        \multicolumn{2}{|c|}{}   & Client 1 & Client 2 & Client 3 & Client 4 & Client 5 & Client 6 & Client 7 & Client 8 & Client 9  & \textbf{Average} \\
        \hline \hline
        \multicolumn{2}{|c|}{\color{red} Local Average ($b_1$ to $b_9$)}
        & 0.76 & 0.75 & 0.71 & 0.72 & 0.67  & 0.70 & 0.76 & 0.64 & 0.82 & \textbf{\zz{0.72}} \\
        \hline
        \multicolumn{2}{|c|}{\color{red} Training Centrally on All Data}
        & 0.87 & 0.87 & 0.77 & 0.80 & 0.75 & 0.77 & 0.82 & 0.70 & 0.92 & \textbf{\zz{0.81}} \\        \hline
        \hline
        \multicolumn{2}{|c|}{\textbf{FedProx}}
        & 0.82 & 0.78 & 0.73 & 0.75 & 0.72 & 0.74 & 0.82 & 0.69 & 0.96 & \textbf{\zz{0.78}} \\
        \hline
        \hline
        \multicolumn{2}{|c|}{FedProx-LG}
        & 0.77 & 0.61 & 0.65 & 0.65 & 0.60 & 0.69 & 0.77 & 0.63 & 0.93 & \textbf{\zz{0.70}} \\
        \hline
        \multicolumn{2}{|c|}{IFCA}
        & 0.83 & 0.79 & 0.73 & 0.76 & 0.71 & 0.75 & 0.82 & 0.69 & 0.87 & \textbf{\zz{0.77}} \\
        \hline
        \multicolumn{2}{|c|}{\textbf{FedProx + Fine-tuning}}
        & 0.84 & 0.89 & 0.79 & 0.78 & 0.72 & 0.75 & 0.82 & 0.72 & 0.90 & \textbf{\zz{0.80}} \\
        \hline
        \multicolumn{2}{|c|}{Assigned Clustering}
        & 0.81 & 0.86 & 0.75 & 0.76 & 0.72 & 0.75 & 0.81 & 0.70 & 0.88 & \textbf{\zz{0.78}} \\
        \hline
        \multicolumn{2}{|c|}{FedProx + $\alpha$-Portion Sync}
        & 0.82 & 0.79 & 0.73 & 0.76 & 0.72 & 0.75 & 0.81 & 0.69 & 0.90 & \textbf{\zz{0.78}} \\
        \hline
    \end{tabular}
    }
\end{table*}

\begin{table*}[ht]
    \centering
     \renewcommand{\arraystretch}{1.1}
     \caption{Testing Accuracy Comparison (ROC AUC) on Routability Prediction with RouteNet~\cite{xie2018routenet}}
     \vspace{-.1in}
     \label{result_routenet}
    \resizebox{\linewidth}{!}{
    \begin{tabular}{| c | c | c | c | c | c | c | c | c | c | c || c | }
        \hline
        \multicolumn{2}{|c|}{}   & \multicolumn{10}{c|}{Testing on}  \\
        \cline{3-12}
        \multicolumn{2}{|c|}{}   & Client 1 & Client 2 & Client 3 & Client 4 & Client 5 & Client 6 & Client 7 & Client 8 & Client 9  & \textbf{Average} \\
        \hline
        \hline
        \multicolumn{2}{|c|}{\color{red} Local Average ($b_1$ to $b_9$)} 
        & 0.76 & 0.76 & 0.71 & 0.73 & 0.68 & 0.71 & 0.75 & 0.64 & 0.78 & \textbf{\zz{0.73}} \\
        \hline        
        \multicolumn{2}{|c|}{\color{red} Training Centrally on All Data}
        & 0.86 & 0.88 & 0.79 & 0.82 & 0.81 & 0.77 & 0.82 & 0.75 & 0.94 & \textbf{\zz{0.83}} \\
        \hline
        \hline
        \multicolumn{2}{|c|}{\textbf{FedProx}}
        & 0.63 & 0.83 & 0.71 & 0.72 & 0.66 & 0.67 & 0.63 & 0.57 & 0.42 & \textbf{\zz{0.65}} \\
        \hline
        \hline
        \multicolumn{2}{|c|}{FedProx-LG}
        & 0.60 & 0.55 & 0.57 & 0.50 & 0.51 & 0.49 & 0.54 & 0.52 & 0.46 & \textbf{\zz{0.53}} \\
        \hline
        \multicolumn{2}{|c|}{IFCA}
        & 0.46 & 0.28 & 0.35 & 0.37 & 0.39 & 0.44 & 0.43 & 0.43 & 0.71 & \textbf{\zz{0.43}} \\
        \hline
        \multicolumn{2}{|c|}{\textbf{FedProx + Fine-tuning}}
        & 0.83 & 0.86 & 0.76 & 0.75 & 0.74  & 0.75 & 0.81 & 0.72 & 0.90 & \textbf{\zz{0.79}} \\
        \hline
        \multicolumn{2}{|c|}{Assigned Clustering}
        & 0.70 & 0.85 & 0.74 & 0.65 & 0.64 & 0.65 & 0.49 & 0.46 & 0.89 & \textbf{\zz{0.67}} \\
        \hline
        \multicolumn{2}{|c|}{FedProx + $\alpha$-Portion Sync}
        & 0.66 & 0.57 & 0.61 & 0.57 & 0.54  & 0.58 & 0.68 & 0.58 & 0.72 & \textbf{\zz{0.61}} \\
        \hline
    \end{tabular}
        }
\end{table*}

\begin{table*}[ht]
    \centering
     \renewcommand{\arraystretch}{1.1}
     \caption{Testing Accuracy Comparison (ROC AUC) on Routability Prediction with PROS~\cite{chen2020pros}}
          \vspace{-.1in}
     \label{result_pros}
    \resizebox{\linewidth}{!}{
    \begin{tabular}{| c | c | c | c | c | c | c | c | c | c | c || c | }
        \hline
        \multicolumn{2}{|c|}{}   & \multicolumn{10}{c|}{Testing on}  \\
        \cline{3-12}
        \multicolumn{2}{|c|}{}   & Client 1 & Client 2 & Client 3 & Client 4 & Client 5 & Client 6 & Client 7 & Client 8 & Client 9  & \textbf{Average} \\
        \hline
        \hline
        \multicolumn{2}{|c|}{\color{red} Local Average ($b_1$ to $b_9$)} 
        & 0.65 & 0.63 & 0.61 & 0.61 & 0.58 & 0.62 & 0.66 & 0.59 & 0.72 & \textbf{\zz{0.63}} \\
        \hline
        \multicolumn{2}{|c|}{\color{red} Training Centrally on All Data}
        & 0.75 & 0.68 & 0.65 & 0.65 & 0.62 & 0.62 & 0.73 & 0.65 & 0.73 & \textbf{\zz{0.67}} \\
        \hline
        \hline
        \multicolumn{2}{|c|}{\textbf{FedProx}}
        & 0.67 & 0.60 & 0.61 & 0.64 & 0.63 & 0.64 & 0.65 & 0.59 & 0.58 & \textbf{\zz{0.62}} \\
        \hline
        \hline
        \multicolumn{2}{|c|}{FedProx-LG}
        & 0.69 & 0.62 & 0.62 & 0.63 & 0.61 & 0.65 & 0.71 & 0.60 & 0.84 & \textbf{\zz{0.66}} \\
        \hline
        \multicolumn{2}{|c|}{IFCA}
        & 0.50 & 0.58 & 0.52 & 0.53 & 0.51 & 0.48 & 0.51 & 0.51 & 0.35 & \textbf{\zz{0.50}} \\
        \hline
        \multicolumn{2}{|c|}{\textbf{FedProx + Fine-tuning}}
        & 0.74 & 0.65 & 0.76 & 0.72 & 0.53 & 0.67 & 0.81 & 0.69 & 0.50 & \textbf{\zz{0.67}} \\
        \hline
        \multicolumn{2}{|c|}{Assigned Clustering}
        & 0.47 & 0.55 & 0.51 & 0.48 & 0.49 & 0.51 & 0.70 & 0.60 & 0.36 & \textbf{\zz{0.52}} \\
        \hline
        \multicolumn{2}{|c|}{FedProx + $\alpha$-Portion Sync}
        & 0.64 & 0.45 & 0.56 & 0.58 & 0.55 & 0.52 & 0.64 & 0.55 & 0.59 & \textbf{\zz{0.56}} \\
        \hline
    \end{tabular}
    }
\end{table*}

\subsection{Experiment Setup}

We construct a comprehensive dataset using 74 designs with largely varying sizes from multiple benchmarks.
There are 29 designs from ISCAS’89~\cite{brglez1989combinational}, 13 designs from ITC'99~\cite{corno2000rt}, 19 other designs from Faraday and OpenCores in the IWLS'05~\cite{albrecht2005iwls}, 13 designs from ISPD'15~\cite{bustany2015ispd}.
For each design, multiple placement solutions are generated with different logic synthesis and physical design settings.
Altogether 7,131 placement solutions are generated from these 74 designs.
We apply Design Compiler\textsuperscript{\textregistered} for logic synthesis and Innovus\textsuperscript{\textregistered}~\cite{Innovus} for physical design, with the NanGate 45nm technology library~\cite{URL:NanGate}.

To validate our algorithm, we mimic a real application scenario by splitting all designs to nine different clients ($K=9$).
Since designs from the same client tend to be more similar to each other, we assign designs from the same benchmark suite to the same client.
Then for each client, we randomly split around $70\%$ of designs to be training data and the other $30\%$ of designs to be the testing data.
Notice that there are no clients sharing common designs, and there are no designs belonging to training and testing data at the same time.
This prevents information leakage between clients and ensures testing designs are completely unseen to trained models.
Details of the design assignment and number of placements in each client are shown in Table~\ref{tbl:designs}.
Three clients collect designs from ITC'99, three clients collect designs from ISCAS'89, two clients collect from Faraday and OpenCores in the IWLS’05, and one client collects from ISPD'15.
For each client, the number of placement solutions in the training data ranges from $175$ to $812$, and the number in the testing data ranges from $84$ to $348$.


We evaluate accuracy with receiver operating characteristic (ROC) area under curve (AUC), measured based on the confusion matrix from prediction. The AUC ranges from 0 to 1, with larger value indicating higher model accuracy. We verify the accuracy in ROC AUC for all proposed federated learning methods with personalization using three models, two representitive routability estimators RouteNet~\cite{xie2018routenet} and PROS~\cite{chen2020pros}, and our proposed FLNet.

For hyperparameters in our experiment, the number of rounds $R=50$, the number of model update steps in each round $S=100$, the number of steps for local fine-tuning $S'=5000$.
We use a learning rate of $0.0002$, Adam optimizer, and an L2 regularization strength of $0.00001$.
The FedProx proximal term strength $\mu$ is $0.0001$.
For the $\alpha$-portion model, we test $\alpha=0.5$.
For FedProx-LG, we set the output layers of the three models to be the local part, while the remaining layers to be the global part.
For IFCA, we set the number of clusters $C=4$.
For the assigned clustering method, we select 4 clusters: Client 1-3, Client 4-6, Client 7-8, and Client 9.

\subsection{Training Method Evaluation}

Table~\ref{result_flnet} shows the accuracy of FLNet, using various model training algorithms based on decentralized private data.
The first baseline is the average performance among $K=9$ locally-trained models $b_1$ to $b_9$, which equals $0.72$.
This corresponds to the performance of traditional ML model construction methods in such a decentralized setting.
In addition, a very strong baseline is training the ML model centrally on all training data.
This indicates a scenario where we can explicitly collect all data together from all clients without data privacy concerns.
Without accuracy degradation caused by heterogeneity among different clients, this accuracy $0.81$ can be viewed as an empirical upper limit we should target for decentralized training.
Ideally, our FL algorithms should achieve the same accuracy if they well address all challenges in the decentralized setting. 

As Table~\ref{result_flnet} shows, our FLNet model trained with FedProx reaches the accuracy of $0.78$, outperforming the locally trained baselines ($b_1$ to $b_9$) by $0.06$ in absolute accuracy value.
When inspecting its performance on each individual client, it outperforms the baseline's performance on every single client.
This FedProx based on FLNet is our proposed method to generate a single generalized model.

To achieve better local accuracy, instead of simply adopting the generalized model from FedProx, we apply different personalization techniques to customize the model for each client.
Among all five different personalization algorithms we proposed, the straightforward FedProx + Fine-tuning achieves the best accuracy in $0.80$, outperforming local models by $0.08$, which is $11\%$ relative improvement.
This accuracy is also very close to the empirical upper limit accuracy ($0.81$) from centralized training.
Compared with simple FedProx, it further improves accuracy at the cost of extra fine-tuning for each client.

\subsection{ML Model Evaluation}
\label{sec:ml-model-evaluation}

The accuracy of representative routability models RouteNet~\cite{xie2018routenet} and PROS~\cite{chen2020pros} are shown in Table~\ref{result_routenet} and Table~\ref{result_pros}, respectively.
Compared with FLNet, RouteNet in Table~\ref{result_routenet} achieves slightly better accuracy for local training and centrally training.
This is reasonable since the RouteNet model is developed under this traditional training setup.
In comparison, for all decentralized training algorithms we proposed, FLNet achieves superior accuracy than RouteNet.
This proves the robustness of our proposed FLNet against client data heterogeneity, thus FLNet is a better choice for federated learning. 

As indicated by Table~\ref{result_routenet}, the FedProx method based on RouteNet is obviously vulnerable to decentralized training and is even less accurate than local models ($b_1$ to $b_9$).
Similarly, most personalization methods are also significantly affected.
Only after finetuning on local clients, those customized models' accuracy rises back to $0.79$.
This is because such finetuning process is no longer under the decentralized setting. 

In Table~\ref{result_pros}, the overall accuracy of PROS, another popular routability model, is much lower than RouteNet and FLNet.
One possible reason is PROS is originally proposed to predict routing congestions, and thus is less effective for DRC violation prediction.
Also, its higher model complexity makes it even more vulnerable to the data heterogeneity.
But we can still observe a trend very similar to RouteNet.
The FedProx model based on PROS is also less accurate than local models ($b_1$ to $b_9$), and after finetuning on local clients, the accuracy rises close to the accuracy of centralized training. 

In summary, the combination of our proposed FLNet and FedProx can well utilize decentralized private data and achieve better performance ($0.78$) than existing methods RouteNet and PROS with the accuracies $0.73$ and $0.63$, respectively.
By further fine-tuning the model on each client, the averaged accuracy of FLNet with FedProx + Fine-tuning rises to $0.80$, which is close to the accuracy of training all data centrally.

%% file: _txt/6_conclusion.tex
\section{Conclusion}

In this work, we bring attention to the serious data availability problem in ML for EDA applications, and propose a federated learning-based solution to encourage data sharing. Our proposed collaborative training method based on FLNet model proves to be robust to the client heterogeneity in decentralized training data. This solution demonstrates its benefit in routability prediction and can be potentially extended to other ML for EDA tasks, especially layout-level predictions and optimizations with similar problem formulations. In the future, as the ML methods get more widely deployed in design flow and the demand on training data keeps increasing, such collaborative training may become a standard training operation. 

%% file: _txt/7_acknowledgements.tex
\section*{Acknowledgments}

This work is supported by NSF CCF-2106725, NSF CCF-2106828, SRC GRC Task 3103.001, and SRC GRC Task 3104.001.